\documentclass[runningheads]{llncs}

\usepackage{eccv}

\usepackage{eccvabbrv}
\definecolor{mypink}{RGB}{255, 100, 203}
\usepackage{graphicx}
\usepackage{booktabs}
\usepackage{wrapfig}

\usepackage[accsupp]{axessibility}  

\usepackage{hyperref}

\usepackage{orcidlink}
\usepackage{multirow}

\begin{document}

\title{Towards Flexible, Natural, Efficient Interaction for Conversational Talking Face Generation} 

\titlerunning{Flexible, Natural, Efficient Interaction for Conversational Talking Face}

\author{Baiqin Wang\inst{1,2} \and Sen Chen\inst{1,2} \and Jiankuo Zhao\inst{1,2} \and Xiangyu Liu\inst{1,2} \and \\  Zhen Lei \inst{1,2,3,4} \and Xiangyu Zhu\inst{1,2}\thanks{Corresponding author.} }
\authorrunning{B.~Wang et al.}

\institute{MAIS, Institute of Automation, Chinese Academy of Sciences \and
	School of Artificial Intelligence, University of Chinese Academy of Sciences \and
	CAIR, HKISI, Chinese Academy of Sciences \quad
	$^{4}$\enspace SCSE, FIE, M.U.S.T\\
	\email{\{wangbaiqin2024, zhen.lei, xiangyu.zhu\}@ia.ac.cn}}

\maketitle

\begin{abstract}
  Conversational talking face generation has recently attracted increasing attention, aiming to synthesize interactive talking videos where characters speak, listen, and respond dynamically to each other. This task presents three core challenges: 1) Flexibility: enabling multi-round dialogues with an arbitrary number of participants; 2) Naturalness: maintaining coherent motion and appropriate non-verbal feedback throughout the interaction; and 3) Efficiency: achieving real-time generation and low computation overhead for long-term continuous online conversation. Despite recent advances, existing methods still fall short in balancing all three requirements. To bridge this gap, we introduce InterTalk, a novel and efficient framework designed for highly interactive conversational talking face generation. Built upon a motion-based architecture, InterTalk supports real‑time conversation synthesis. Our method achieves strong flexibility by explicitly modeling multi‑round conversational dynamics among each participant, eliminating constraints on their numbers. To enhance interactivity, we incorporate motion feedback from multiple participants and introduce an iterative generation strategy for more natural behaviors. Besides, we disentangle motion into several facial components, enabling targeted refinements for natural response such as precise lip‑sync and realistic eye‑blinking. Finally, we construct a new multi‑person conversational dataset and enrich it with 3D face‑based data augmentation. Extensive experiments demonstrate that InterTalk achieves superior interaction quality while maintaining real-time performance at 30 FPS. Project page:	 \textcolor{mypink}{ https://bq-wang0511.github.io/InterTalk/}
\end{abstract}

\begin{figure*}[t] 
    \centering
    \includegraphics[width=\linewidth]{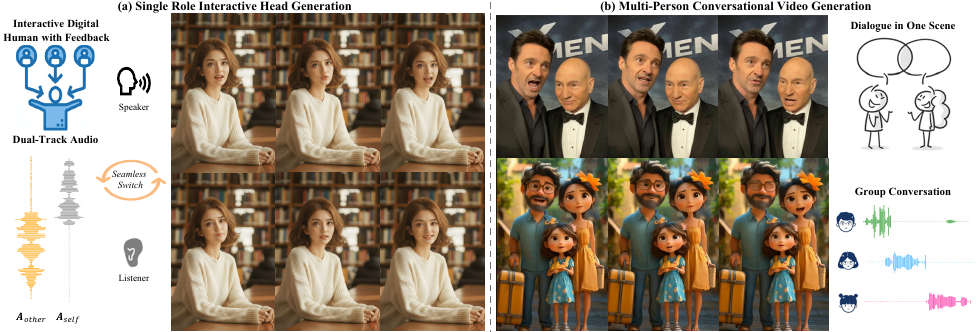}
    \vspace{-5mm}
    \caption{\textbf{Illustration of InterTalk.}
   InterTalk generates conversational talking face with flexible and natural interactions in real-time. (a) For a single participant, it generates an interactive talking face capable of multi‑round dialogues with seamless role switching. (b) For multiple participants, it produces coherent talking‑face videos within a shared scene, capturing realistic group interactions.}
    \label{fig:first_pic}
    \vspace{-5mm}
\end{figure*}

\section{Introduction}
\label{sec:intro}

Talking face generation aims to create expressive facial animations with lip movements synchronized to speech. While existing methods~\cite{zhang2023sadtalker,chen2025echomimic,ji2025sonic,wang2025pc,guo2024liveportrait} mainly focus on only speaking scenarios, there remains great potential in interactive conversational talking faces between participants~\cite{zhu2025infp,guo2025arig,kong2025let}. Unlike single-speaker generation, this task emphasizes fluid interactions among characters, including non-verbal feedback such as nods or subtle expression changes. Realistic conversational agents can significantly enhance user engagement in applications such as online education, remote meetings and emotionally digital companions.

In this paper, we identify three key requirements for conversational talking face generation. \textit{Flexibility} demands the ability to model multi-round conversations between numerous participants. \textit{Naturalness} calls for coherent interactions, including smooth articulation by speakers and timely non-verbal feedback from listeners, along with seamless transitions between characters. \textit{Efficiency} demands real-time performance with low computational cost, which is critical for long-term continues applications like virtual companion. These aspects collectively form the foundation for advancing conversational talking face generation.

Current approaches for conversational talking face generation can be broadly divided into two categories. The first category explicitly models conversational dynamics to improve efficiency and interactivity but still faces several limitations. Some methods~\cite{tran2024dim,wang2024disentangling} generate speaker and listener motions separately, requiring manual role switching which hinders multi-round conversations, while others, like INFP~\cite{zhu2025infp}, rely on dual-track audio but focus on single-character scenes without feedback from others, resulting in isolated motions and weak responsiveness, confined to single‑role scenarios. Additionally, their lip-sync accuracy remains restricted by limited conversational data. The second leverages video generation models~\cite{blattmann2023stable,peebles2023scalable,wan2025wan} to synthesize multi-person conversation videos in an end-to-end manner, such as MultiTalk~\cite{kong2025let}, which adopts Wan 2.1~\cite{wan2025wan} as its backbone. However, these methods are computationally intensive and rely on extremely large-scale training datasets, which is not suitable for application that requires high throughput generation including companion digital human and live-streaming.  In summary, existing works struggle to simultaneously fulfill the three essential requirements of flexibility, naturalness, and efficiency.
 
To achieve these objectives, we propose InterTalk, a framework designed for both single-role and multi-person conversational scenarios as shown in Fig.~\ref{fig:first_pic}. For the single-role interactive head generation, InterTalk provides coherent non-verbal feedback and seamlessly transitions between speaking and listening states. When it comes to multi-person scenarios, our method supports dialogue generation for an arbitrary number of participants within a shared scene for group conversation. Built on a paradigm that explicitly models interaction dynamics, InterTalk first incorporates a \textbf{Responsive Context Encoder} (RCE) to integrate all participants’ motion and audio cues into a unified interactive latent representation. This representation is then utilized by the \textbf{Interactive Motion Generator} (IMG) to produce \textit{Flexible} multi-round conversational motions for each participant. To enhance interactivity, our framework incorporates motion feedback from all participants and adopts an \textbf{Iterative Generation Strategy} that progressively refines each participant’s motion, ensuring \textit{Natural} and lifelike multi‑party interactions. Moreover, as different facial components exhibit distinct behavioral patterns—for instance, lip motion closely correlates with audio while eye movement is more constrained—we \textbf{Disentangle Facial Motion} into several components and model them separately. This design yields more natural behaviors and provides fine‑grained control, where lip‑sync accuracy is enhanced by a sync‑enhancer module trained on single‑speaker data, and an eye‑blinking enhancer together with a head‑pose initializer further improve visual realism. With an efficient  rendering pipeline, InterTalk synthesizes multi-participant talking videos in \textit{Real-Time} with low computation overhead.

Given the absence of publicly available interaction data involving multiple participants, we introduce a newly collected multi-person talking face dataset to bridge this gap. To further enhance the diversity and scale of the training data, we employ a \textbf{3D Face-based Data Augmentation} that converts 3D FLAME~\cite{FLAME:SiggraphAsia2017} coefficients into implicit keypoint, which provides additional supervisory signals on motions during model training.

In summary, our contributions are as follows:
\begin{itemize}
\item  We propose \textbf{InterTalk}, a novel and efficient framework that  generating multi‑round conversational videos with an arbitrary number of participants, which simultaneously satisfies three key requirements of  \textbf{Flexibility},  \textbf{Naturalness}, and  \textbf{Efficiency} in conversational talking face generation.
\item To enhance the interactivity, we incorporate motion feedback and introduce an \textbf{Iterative Generation Strategy} that enables each character’s motion to faithfully reflect feedback from others. We also \textbf{Disentangle Facial Motion} into distinct components, leveraging their distinct patterns to achieve more natural interactions and fine‑grained controllability.
\item We construct a new multi‑person conversational dataset and further enhance training through a  \textbf{3D Face-based Data Augmentation}, which converts 3D facial coefficients into implicit keypoints for effective supervision.
\item Extensive experiments demonstrate that InterTalk achieves state‑of‑the‑art performance in conversational talking face generation, delivering flexible, natural, and efficient interactions between participants.
\end{itemize}

\section{Related Work}
\label{sec:related_work}

\subsection{Single-person Talking Face Generation}

Talking‑face generation has made remarkable progress in recent years. Early methods such as Wav2Lip~\cite{prajwal2020lip} adopt GAN‑based architectures~\cite{goodfellow2014generative} to synchronize lip motion with audio, but only modify the mouth region while keeping head pose and expressions fixed. More recently, several approaches~\cite{ji2025sonic,chen2025echomimic,cui2024hallo2} leverage powerful video diffusion models to synthesize temporally coherent and fluid talking videos, yet their heavy computational cost limits real‑time deployment. Apart from them, several works ~\cite{zhang2023sadtalker,wang2025pc} adopt a two‑stage generation strategy, first producing an intermediate representation—such as facial landmarks or implicit keypoints ~\cite{guo2024liveportrait} —and then rendering the final output conditioned on an identity image. This design achieves a strong balance between efficiency and flexibility.

In addition to speaking-oriented methods, a number of approaches focus on listening‑head generation, which aims to produce non‑verbal reactions—such as head nods or smiling expressions—driven by the speech of other speakers. For example, L2L~\cite{ng2022learning} employs a codebook‑based model to generate listening motions, while ELP~\cite{song2023emotional} incorporates emotional cues derived from speech content. DIM~\cite{tran2024dim} can transition between talking and listening states but cannot generate both states simultaneously, which limits its applicability for interactive head generation.

\subsection{Conversational Talking Face Generation}

Unlike single‑person scenarios, conversational talking face generation focuses on modeling interactions among participants. Differing from early methods~\cite{ng2022learning,song2023emotional,tran2024dim}  that rely on manual role switching between speakers and listeners, INFP~\cite{zhu2025infp} is the first approach to generate interactive talking videos from dual audio streams, synthesizing motions conditioned on both speech and feedback without explicit role switching. ARIG~\cite{guo2025arig} further extends it through an auto‑regressive model. While these methods enable long multi‑turn dialogues, they still suffer from limited partner‑feedback modeling and remain constrained to single‑person animation. Some 3D facial animation approaches, such as DualTalk~\cite{peng2025dualtalk} also attempt to model dialogues with motion feedback, but their mesh‑based generation restricts their applicability to 2D video synthesis. 

For multi‑person video animation, recent works increasingly employ video generation models to synthesize multi‑character conversations. For example, MultiTalk~\cite{kong2025let} adopts Wan 2.1~\cite{wan2025wan} as its backbone with a human‑binding strategy to produce multi‑person animations, but it cannot scale beyond two participants. Emerging systems such as Veo3~\cite{google2025veo3} and Sora2~\cite{openai2024sora} achieve impressive audio‑visual joint generation but are computationally costly, remain closed sourced, and require massive data, which cannot be adopted for high-throughput application like companion. In contrast, our method explicitly models interactions among multi-participants and achieves efficient real‑time animation.

\section{Method}
\label{sec:method}

\begin{figure*}[t!]
    \centering
    \includegraphics[width=\textwidth]{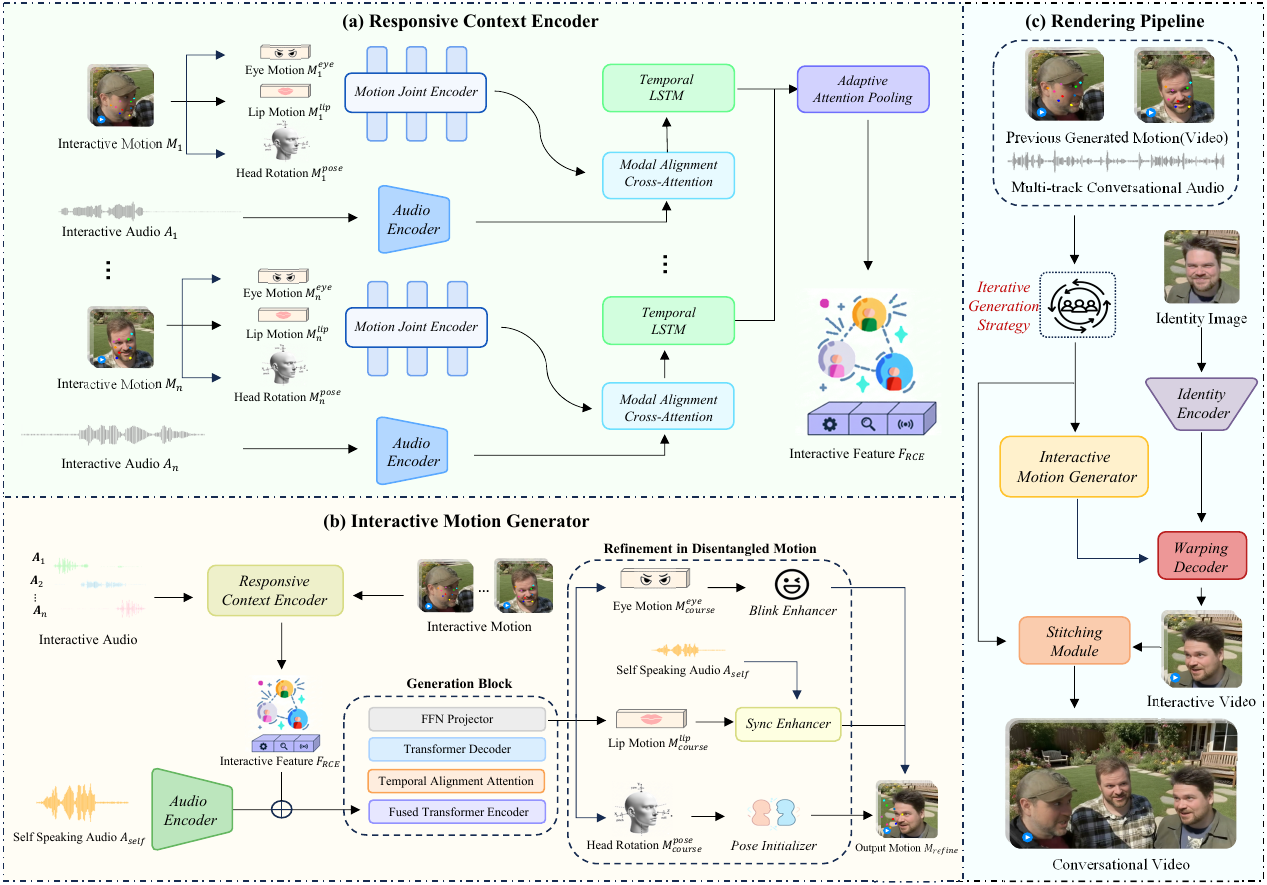}
      \vspace{-3mm} 
    \caption{\textbf{Framework of InterTalk.}
InterTalk consists of three key components: a Responsive Context Encoder (RCE) that integrates environmental elements into an interactive feature representation; an Interactive Motion Generator (IMG) that produces fluid conversational motions; and a Rendering Pipeline that animates each participant to synthesise the final talking‑face videos.   }
    \label{fig:framework}
    \vspace{-3mm} 
\end{figure*}

Our proposed framework InterTalk aims to generate conversational talking face that achieve flexibility, naturalness, and efficiency simultaneously. As illustrated in Fig.~\ref{fig:framework}, we first apply a Responsive Context Encoder (RCE) to captures interactive cues from the environment, including the speech and motion feedback of other participants during conversation. An Interactive Motion Generator (IMG) then leverages these interaction features together with the self audio to generate corresponding motions. Finally, the Rendering Pipeline synthesizes talking videos for each participant based on the generated motions and composites them into a cohesive multi‑person conversational scene. Each component plays a vital role and their detailed designs are described in the following sections.

\subsection{Responsive Context Encoder}

The RCE integrates environmental information into a highly interactive representation, providing feature $F_{\text{RCE}}$ for motion generation in the IMG. 
Its overall process can be expressed as:
\begin{equation}
    F_{\text{RCE}} = 
    \text{RCE}\!\left(
        \{A_i, M_i\}_{i=1}^{N}
    \right),
    \label{eq:rce_overall}
\end{equation}
where $A_i$ and $M_i$ denote the audio and motion inputs of the $i$‑th participant, and $N$ is the total number of participants.  For each participant $i$, this module encodes both audio and motion inputs into a unified latent space. Specifically, the audio stream of participant $i$ is encoded using Wav2Vec 2.0~\cite{baevski2020wav2vec} to obtain a robust speech representation $F_{\text{a}}^{i}$.  Meanwhile, the motion input is disentangled into several facial components—lip motion $M_{i}^{\text{lip}}$, eyes motion $M_{i}^{\text{eye}}$, and head pose motion $M_{i}^{\text{pose}}$ —each encoded separately and then combined into a comprehensive motion representation $F_{\text{m}}^{i}$ to better capture non‑verbal feedback within conversations. 
The encoding process is defined as:
\begin{equation}
    F_{\text{a}}^{i} = \text{Enc}_{\text{a}}(A_i), \quad
    F_{\text{m}}^{i} = \text{Enc}_{\text{m}}(M_{i}^{\text{lip}}, M_{i}^{\text{eye}}, M_{i}^{\text{pose}}).
    \label{eq:rce_encode}
\end{equation}

After obtaining the audio and motion features of each participant, we apply a cross‑attention mechanism to align and fuse them across modalities, bridging the gap between the two feature domains. To further enhance temporal coherence, a bi‑directional LSTM~\cite{sherstinsky2020fundamentals} is subsequently applied to capture both past and future conversational dynamics to get $i$‑th participants interactive context feature $F_{\text{RCE}}^{i}$. The overall process for the $i$‑th participant is formulated as:
\begin{equation}
    F_{\text{RCE}}^{i} = \text{biLSTM}\!\left(\text{CrossAttn}(F_{\text{a}}^{i}, F_{\text{m}}^{i})\right).
    \label{eq:rce_fuse_bilstm}
\end{equation}

Note that the context window covers only about 3 seconds rather than a significant future buffer, which is suitable for practical deployment. Finally, the RCE aggregates interactive features across all participants via adaptive attention pooling. Specifically, the reference feature \(\bar{F}=\frac{1}{N}\sum_{i=1}^{N}F_{\text{RCE}}^{i}\) is obtained by averaging all participant features, while attention weights are computed based on each participant’s similarity to this reference and dimension $d$ of $F_{\text{RCE}}^{i}$, yielding higher weights for features closer to the conversational context. The process can be formulated as:
\begin{equation}
    \begin{aligned}
        \alpha_i = \text{Softmax}\!\left(\frac{{F_{\text{RCE}}^{i}}^{\!\top} \bar{F}}{\sqrt{d}}\right), 
        F_{\text{RCE}} = \sum_{i=1}^{N} \alpha_i F_{\text{RCE}}^{i}.
    \end{aligned}
    \label{eq:rce_attention}
\end{equation}
 
 In this way, the RCE produces a unified interactive context feature $F_{\text{RCE}}$ that effectively captures cross‑participant dependencies and provides global conversational context for downstream motion generation.

 \subsection{Interactive Motion Generator}

After obtaining interactive features $F_{\text{RCE}}$ from the RCE, the IMG generates temporally coherent and context‑aware facial motions conditioned on both the self speaker’s audio and environmental interaction cues. The overall process can be formulated as:
\begin{equation}
    M_{\text{refine}} = 
    \text{IMG}\!\left(A_{\text{self}}, F_{\text{RCE}}\right),
    \label{eq:img_overall}
\end{equation}
where $A_{\text{self}}$ denotes the self speaking audio, which is encoded into $F_{a}$ using the same audio encoder as in the RCE to ensure feature‑space consistency.    The  $F_{a}$ is combined with the interactive feature $F_{\text{RCE}}$ from Eqn.~\ref{eq:rce_overall} before being fed into the motion generation network to produce $M_{\text{coarse}}$. It can be formulated as:
\begin{equation}
    M_{\text{coarse}} = 
    \text{GenBlock}\!\left(F_{\text{RCE}} + F_{a}\right).
    \label{eq:img_genblock}
\end{equation}

The generation network comprises four sequential blocks. A fused Transformer Encoder~\cite{vaswani2017attention} first integrates audio and interactive features within a unified latent space. Next, a temporal alignment attention module, adapted from FaceFormer~\cite{fan2022faceformer}, applies a causal mask to ensure temporal consistency and prevent future‑frame leakage.  A Transformer decoder then captures long‑range dependencies and contextual transitions, followed by a feed‑forward layer that projects the learned representations into subspaces of different facial components—lip, eye, and head pose—for targeted motion refinement.

\noindent\textbf{Refinement on Disentangled Motion.} We apply three modules to refine the decomposed motion from the coarse motion $M_{\text{coarse}}$. To improve lip‑sync accuracy beyond the limitations of conversational data, we train a lightweight sync-enhancer on a large‑scale single‑speaker dataset. This module takes speech audio $A_{\text{self}}$ and initial lip motion $M_{\text{coarse}}^{\text{lip}}$ as input and predicts local deformations for refined synchronization. It is trained independently after other modules are pretrained, ensuring compatibility with the full system. We further incorporate a Blink‑Enhancer that regulates blink frequency $f$ and noise $z$ to produce natural eyelid closures. In addition, a Pose‑Initializer adjusts the initial head orientation according to the relative facial position $(x_{\text{self}}, y_{\text{self}})$, maintaining mutual gaze alignment at the start of the interaction. The refinement process can be formulated as:
\begin{equation}
\left\{
\begin{aligned}
    M_{\text{refine}}^{\text{eye}}  &= \text{BlinkEnhancer}(M_{\text{coarse}}^{\text{eye}}, f, z), \\
    M_{\text{refine}}^{\text{lip}}  &= \text{SyncEnhancer}(M_{\text{coarse}}^{\text{lip}}, A_{\text{self}}), \\
    M_{\text{refine}}^{\text{pose}} &= \text{PoseInitializer}(M_{\text{coarse}}^{\text{pose}}, x_{\text{self}}, y_{\text{self}}).
\end{aligned}
\right.
\label{eq:disentangle}
\end{equation}

We then combine these refined components to obtain the final output $M_\text{refine}$. Further details on the target‑specific enhancement are provided in the supplementary material.


\subsection{Rendering Pipeline}

While most prior approaches focus on generating a single speaker~\cite{zhang2023sadtalker,wang2025pc,zhu2025infp,guo2025arig}, our rendering pipeline synthesizes coherent conversational videos with multiple participants. We adopt 3D implicit keypoints as intermediate representations, extracted through three modules: a pose estimator computing rotation $R$, translation $t$, and scale $s$; an expression estimator predicting deformation $\delta$; and a canonical keypoint detector extracting canonical keypoints $K_c$ from reference images $I_{ref}$. The original keypoints $K_{ori}$ are obtained as:
\begin{equation}
    \begin{aligned}
        K_{ori} = s \cdot (K_c \cdot R + \delta) + t.
    \end{aligned}
    \label{eq:transform}
\end{equation}

The refined motion $M_{\text{refine}}$ is decomposed into rotation $R_d$ and expression deformation $\delta_d$, which are then applied to transform $K_{ori}$ into the driven implicit keypoints $K_d$. The process can be formulated as:
\begin{equation}
    \begin{aligned}
        M_{\text{refine}} \Rightarrow \{R_d, \delta_d\}, 
        K_{d} = s \cdot (K_c \cdot R_d + \delta_d) + t.
    \end{aligned}
    \label{eq:transform_kd}
\end{equation}

For each participant, the region is cropped from $I_{ref}$ and encoded into an appearance embedding $f_a$ via an identity encoder. 
A warping decoder fuses $f_a$ with the motion sequence $K_d$ predicted by the IMG to estimate a dense warping flow field. This flow is then applied to $f_a$ to generate the final rendered frame with accurate alignment and natural motion:
\begin{equation}
    I_{res} = \text{Decoder}\!\left(\text{Warp}(f_a, K_{ori}, K_d)\right).
    \label{eq:warp}
\end{equation}

Finally, we introduce a stitching module that spatially aligns and blends individual renderings. It refines local misalignment via fine‑grained keypoint deformations and uses a spatial blending mask for smooth transitions. This allows InterTalk to seamlessly composite all animated participants into the original scene, producing visually coherent multi‑participant animations. Further details are provided in the supplementary material.

\subsection{Training and Inference}

\noindent\textbf{Loss Function.} 
We use the L2 loss $\mathcal{L}_{\text{rec}}$ and velocity loss $\mathcal{L}_{\text{vel}}$ as supervision on motion, which can be formulated as:
\begin{equation}
    \mathcal{L}_{\text{motion}} = 
    \lambda_{\text{rec}} \mathcal{L}_{\text{rec}} + 
    \lambda_{\text{vel}} \mathcal{L}_{\text{vel}}.
    \label{eq:loss_motion}
\end{equation}
This loss ensures motion fidelity and temporal smoothness across frames. In addition, the sync‑enhancer is trained on single‑person talking‑face data with a synchronization loss $\mathcal{L}_{\text{sync}}$~\cite{prajwal2020lip} to refine the lip–audio alignment between speech and facial motion. Further training details and explanations are provided in the supplementary material.

\begin{figure*}[t!]
 	\centering
 	\includegraphics[width=0.75\linewidth]{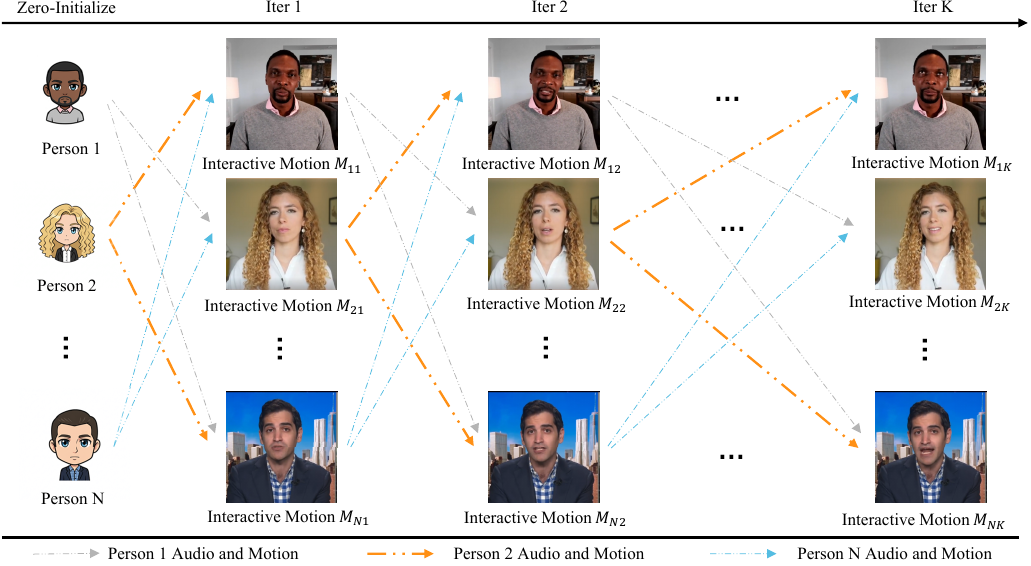}
 	\vspace{-2mm}
 	\caption{\textbf{Iterative Generation Strategy.}
 		By progressively updating motions through multiple iterations, 
 		the final results accurately capture mutual feedback among all participants.}
 	\label{fig:iteration}
 	\vspace{-4mm} 
 \end{figure*}

\noindent\textbf{Iterative Generation Strategy.} 
During inference, the model can only utilize motion feedback from previously generated participants rather than all individuals, which may compromise interactive dynamics. To mitigate this limitation, we introduce an iterative motion generation strategy that progressively incorporates feedback across cycles. As illustrated in Fig.~\ref{fig:iteration}, we sequentially generate motions by using the outputs from the previous iteration as inputs for the current one, continuously updating a shared motion buffer. Through several iterations, each participant’s motion gradually incorporates feedback from others, 
resulting in more responsive and natural interactions without delay.  Formally, let $M_{ij}$ denote the motion of the $i$-th participant at iteration $j$.  The iterative process can be defined as:
\begin{equation}
    M_{ij} = \text{IMG}\!\left(A_i,\, \text{RCE}\!\left(M_{(j-1)}^{\neg i}, A_{(j-1)}^{\neg i}\right)\right),
    \label{eq:feedback}
\end{equation}
where $A_i$ represents the audio feature of the $i$-th participant, and RCE encodes the interaction context from the previous iteration's motions $M_{(j-1)}^{\neg i}$ and audio features $A_{(j-1)}^{\neg i}$ of all other participants except themselves ($\neg i$). The initial feedback motions $M_{i0}$ and audio features $A_{0}^{\neg i}$ are zero-initialized at the start of inference. During the iterative process, the three target-refinement modules in IMG are disabled. Since the IMG is lightweight and computationally efficient, this strategy introduces negligible overhead compared with the overall runtime, preserving real-time generation capability. 

\noindent\textbf{Video Dubbing and Single-role Generation.}
Our method can be extended to video inputs by processing each frame as \(K_{ori}\), enabling video dubbing with interactive conversations while preserving natural body motion and background changes. For single‑role interactive head generation, we disable pose initialization and iterative generation strategy, directly animating from multi‑track audio and motion feedback when available.

\section{Data Construction}
\label{sec:data}
Existing public datasets for interactive talking faces remain limited in quality and conversational depth. For instance, ViCo~\cite{zhou2022responsive} focuses on single‑round dialog, while DualTalk~\cite{peng2025dualtalk} provides only 3D FLAME~\cite{FLAME:SiggraphAsia2017} coefficients. Moreover, most multi‑person datasets concatenate separate single‑speaker recordings instead of capturing participants in a shared scene. To overcome these issues, we build a new multi‑person conversational dataset and enrich it with 3D data by converting FLAME coefficients into motions for our framework.

\begin{figure*}[t!]
	\centering
	\includegraphics[width=\textwidth]{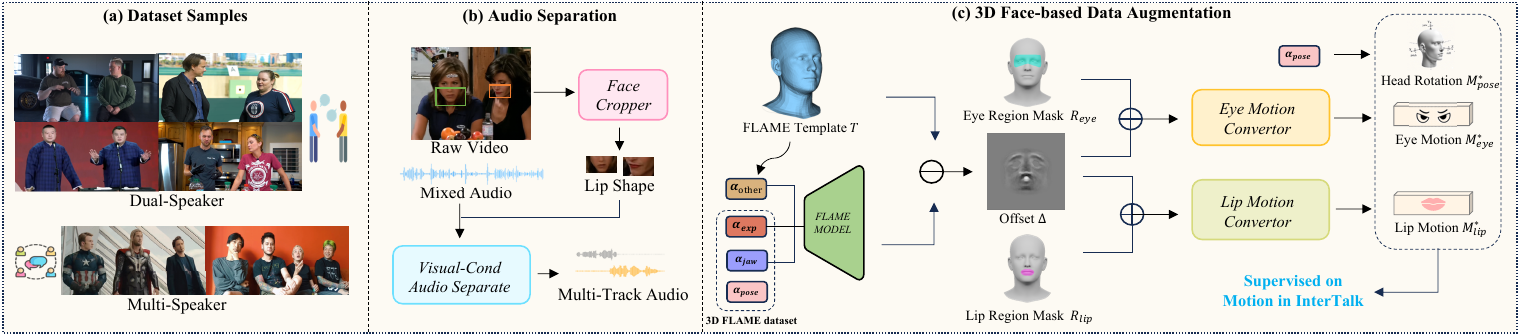}
	\vspace{-4mm}
	\caption{\textbf{Dataset Information.}
		(a) Samples from InterTalk dataset.(b) The processing pipeline for collected videos and multi‑track audio separation.
		(c) 3D face-based data augmentation strategy using 3D conversation datasets, generating motions from FLAME coefficients.}
	\label{fig:dataset}
	\vspace{-3mm} 
\end{figure*}

\noindent\textbf{2D Data Collection.}
As shown in Fig.~\ref{fig:dataset}(a), we collect multi‑speaker conversation videos from the Internet and manually segment them to ensure that each clip includes at least two participants continuously appearing in the same scene. After cleaning, the dataset contains 143 identities and 205 clips. To separate individual audio tracks from mixed recordings, we employ a visual‑audio separation network~\cite{li2024iianet} that leverages lip movements to infer each speaker’s voice as illustrated in Fig.~\ref{fig:dataset}(b). Additional TTS‑based audio segments are synthesized to enhance diversity during evaluation.

\noindent\textbf{3D Face-based Data Augmentation.}
To further enhance conversational diversity and interactive learning, we propose a 3D face-based augmentation strategy that integrates DualTalk’s~\cite{peng2025dualtalk} 3D face data into training. As shown in Fig.~\ref{fig:dataset}(c), the 3D FLAME~\cite{FLAME:SiggraphAsia2017} coefficients are converted into motion signals through two converter networks trained on a 2D talking‑face dataset~\cite{zhang2021flow}, where the FLAME parameters are annotated using EMOCA~\cite{danvevcek2022emoca}. Specifically, we reconstruct a 3D mesh using the jaw pose $\alpha_{\text{jaw}}$ and expression $\alpha_{\text{exp}}$, while keeping the other FLAME coefficients $\alpha_{\text{other}}$ identical to the template mesh $T$. The vertex offsets $\Delta$ are then obtained by subtraction, formulated as:
\begin{equation}
\Delta = \text{FLAME}(\alpha_{\text{exp}},\alpha_{\text{jaw}},\alpha_{\text{other}}) -T.
\end{equation}

Afterward, we apply eye and lip region masks $R_{\text{eye}}$ and $R_{\text{lip}}$ to extract localized expressions, which are fed into corresponding converters to generate motion features as:
\begin{equation}
M_{c}^{*} = \text{Convertor}_{c}\bigl(R_{c}(\Delta)\bigr), \quad c \in \{\text{eye}, \text{lip}\}.
\end{equation}

Head‑pose information is directly mapped to the motion $M_{\text{pose}}^{*}$. This converted dataset further enhances the realism and fluidity of conversational behavior, improving both inter‑participant responsiveness and motion coherence.

 \begin{table}[b]
\setlength{\tabcolsep}{4pt}
\footnotesize
\centering
\vspace{-4mm}
\caption{Quantitative results with state-of-the-art talking face generation methods on HDTF\cite{zhang2021flow} dataset.}
\vspace{-3mm}
\resizebox{0.7\linewidth}{!}{
\begin{tabular}{c ccccc}
\toprule
    Method & Sync-C $\uparrow$ & Sync-D $\downarrow$ & FID $\downarrow$ & NIQE $\downarrow$ & FVD $\downarrow$ \\ \midrule
    SadTalker\cite{zhang2023sadtalker}  &  7.15  & 7.93  & 40.75  &  46.30 &291.66\\
    Hallo2\cite{cui2024hallo2} &  7.53  & 7.97  &  33.15 & 14.25 & 205.60\\
    EchoMimic\cite{chen2025echomimic} &  5.94  &  9.11  &  \underline{28.13} &\underline{ 13.42 }& 284.38\\
    Sonic\cite{ji2025sonic} & \underline{8.26}  &  \textbf{6.88}  &  34.41 &  13.88 & \underline{204.55}\\
    \midrule
    \textbf{Ours} &  \textbf{8.30}  &  \underline{6.91}  &  \textbf{23.07}  & \textbf{13.29} & \textbf{178.73} \\
      
\bottomrule
\end{tabular}
}

\label{table:talking_head_cmp}
\end{table}
\section{Experiments}
\label{sec:experiments}

\subsection{Experimental Settings }

\noindent\textbf{Implementation Details.} 
We preprocess the dataset by converting videos to 25 fps and resampling audio at 16kHz. The RCE and IMG modules are trained using the Adam optimizer~\cite{kingma2014adam} with a learning rate of \(1 \times 10^{-4}\). The components of the rendering pipleline are initialized from LivePortrait~\cite{guo2024liveportrait} to leverage its robust rendering quality. Training is performed on eight RTX 4090 GPUs for three days. During inference, each participant is animated in parallel across multiple GPUs to achieve faster generation speed. Our method produces 512{\texttimes}512 resolution outputs before stitching them back into the original frame. Additional implementation details are provided in the supplementary material.

\noindent\textbf{Metrics.} We evaluate our method across various scenarios, covering both speaking and interactive aspects. For speaking evaluation, we assess lip synchrony using Sync‑C and Sync‑D from SyncNet~\cite{Chung_2016_syncnet}, image quality with FID~\cite{Seitzer2020FID} and NIQE~\cite{mittal2012making} for reference and no-reference assessment, and temporal consistency with FVD~\cite{unterthiner2019fvd}. For interactive evaluation, we extract 3D FLAME~\cite{FLAME:SiggraphAsia2017} coefficients from generated videos and perform quantitative analysis using multiple metrics, including Fréchet Distance (FD), Paired Fréchet Distance (P‑FD), Mean Squared Error (MSE), Similarity Index for Diversity (SID), and Residual Pearson Correlation Coefficient (rPCC). We further report generation speed in frames per second (fps) to measure system efficiency.
\subsection{Comparison}

 \begin{figure}[t!]
	\centering
	\includegraphics[width=\linewidth]{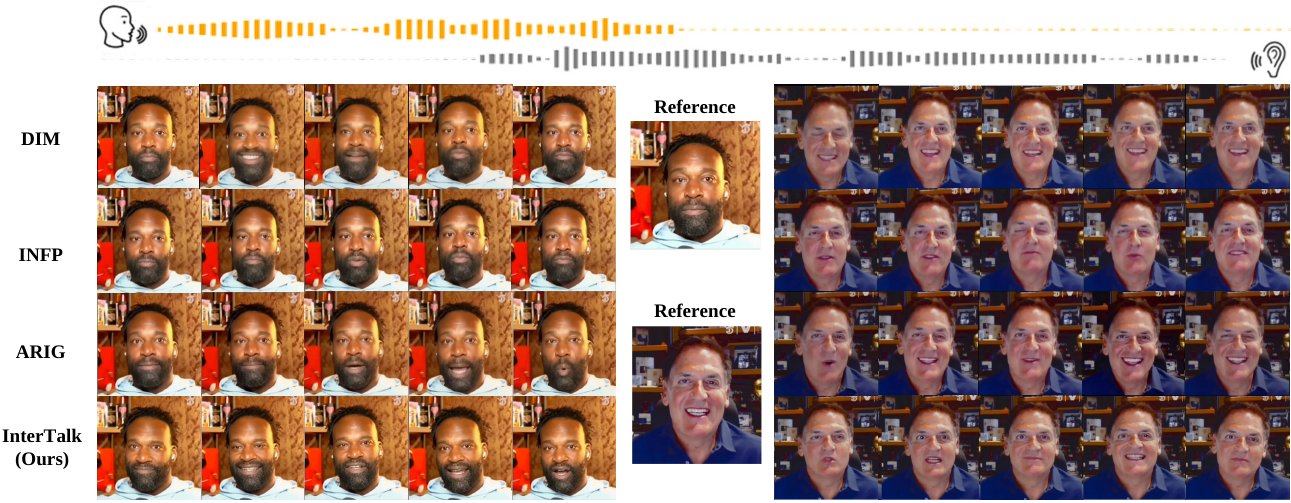}
	\vspace{-3mm} 
	\caption{\textbf{Comparison of Interactive Head Generation.} Two identities are sampled from the DyConv~\cite{zhu2025infp} dataset proposed by INFP, which is not publicly available and therefore lacks ground‑truth results.}
	\label{fig:listen}
	\vspace{-3mm} 
\end{figure}

\noindent\textbf{Talking Face Generation.} 
We first evaluate our method on general single person talking face generation using metrics that assess image quality and lip synchronization. To adapt our framework on this task, we set the interactive audio and motion feedback inputs to zero. We compare our approach with four advanced baselines, including SadTalker~\cite{zhang2023sadtalker}, EchoMimic~\cite{chen2025echomimic}, Hallo2~\cite{cui2024hallo2}, and Sonic~\cite{ji2025sonic} on the HDTF~\cite{zhang2021flow} dataset. As shown in Tab.~\ref{table:talking_head_cmp}, our method demonstrates superior performance in both video quality and lip synchronization. We achieve better FID and NIQE scores owing to the robust rendering pipeline, and obtain the best SyncNet score, slightly outperforming Sonic thanks to the integration of the sync‑enhancer within the IMG. The higher FVD score further indicates that our method achieves better temporal consistency. The qualitative results are demonstrated in supplementary.

 \begin{figure}[t!]
	\centering
	\includegraphics[width=\linewidth]{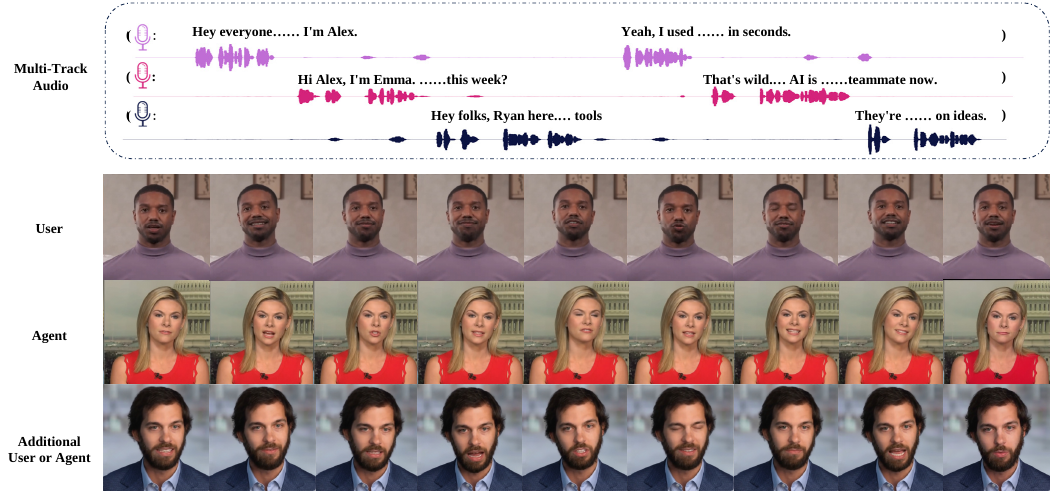}
	\vspace{-4mm}
	\caption{\textbf{Interactive Head Generation.} Our method generates highly responsive interactive agent avatars and supports additional user motion inputs or agent outputs from multi‑track audio, as InterTalk handles an arbitrary number of participants.}
	\label{fig:single}
	\vspace{-3mm} 
\end{figure}

 \begin{table}[b]
\footnotesize
\centering
\vspace{-2mm}
\setlength\tabcolsep{0.15cm}
\caption{Quantitative results with state-of-the-art interactive head generation methods on ViCo\cite{zhou2022responsive} dataset.}
\vspace{-1mm}
\resizebox{0.8\linewidth}{!}{
\begin{tabular}{c cccccc cc}
\toprule
    \multirow{2}[1]{*}{Methods} &\multicolumn{2}{c}{FD $\downarrow$} &   \multicolumn{2}{c}{RPCC $\downarrow$} & 
  \multicolumn{2}{c}{SID$\uparrow$} & \multicolumn{2}{c}{MSE$\downarrow$} \\ 

  \cmidrule(lr){2-3} \cmidrule(lr){4-5} \cmidrule(lr){6-7}
  \cmidrule(lr){8-9}

    &  exp  &  pose &  exp  &  pose &  exp  &  pose &  exp &  pose \\  
    \midrule
    DIM\cite{tran2024dim} &  23.88 &  0.06 &  0.06 & 0.03 & 3.71 & 2.35 & 0.70 & 0.02 \\
    INFP\cite{zhu2025infp} & 18.63 & 0.07 & - & - & 4.78 & 3.92 & \underline{0.51} & \textbf{0.01} \\
    ARIG\cite{guo2025arig} &   \underline{18.39} & 0.05 & \textbf{0.05} & \textbf{0.01} & \underline{4.82} & \underline{3.94} & - & - \\
    DualTalk\cite{peng2025dualtalk} &   22.27 & \textbf{0.04} & 0.07 & 0.03 & 4.36 & 3.84 & 0.58 & \textbf{0.01} \\
    \midrule
    \textbf{Ours} &  \textbf{18.33} &  \textbf{0.04} &  \textbf{0.05} &  \textbf{0.01}  &  \textbf{4.85} &  \textbf{4.07} &  \textbf{0.48}  &  \textbf{0.01} \\

\bottomrule
\end{tabular}
}
\label{table:listening_head_cmp}
\end{table}
 
\noindent\textbf{Interactive Head Generation.} 
We evaluate our method against existing other interactive head generation approaches, including DIM~\cite{tran2024dim}, INFP~\cite{zhu2025infp}, and ARIG~\cite{guo2025arig}. For quantitative evaluation, we conduct experiments on the ViCo~\cite{zhou2022responsive} dataset for listening‑head generation and adopt the metrics reported in the original papers due to the lack of official implementations. As shown in Tab.~\ref{table:listening_head_cmp}, our method achieves superior accuracy and diversity, demonstrating its clear advantage in interactive head generation. In terms of visual results, since INFP and ARIG have not released their source code, we compare our results using demo videos available on their project pages. As shown in Fig.~\ref{fig:listen}, our approach produces natural and expressive head movements with sharper facial details and better identity preservation. In contrast, DIM and INFP display relatively rigid expressions, while ARIG maintains an almost static head pose. In comparison, our method generates more context‑aware gestures—such as gentle nods and natural smiles——demonstrating strong responsiveness during interaction. Furthermore, as shown in Fig.~\ref{fig:single}, our method generates lifelike virtual agents that respond naturally to users. Since it supports an arbitrary number of participants, both the number of output agents and input users are unrestricted, enabling applications like multi-participants remote meeting.

\noindent\textbf{Conversations with Multi-Participants.}
 Due to the scarcity of publicly available multi‑person conversation video and related methods, we evaluate our approach on our collected dataset with MultiTalk~\cite{kong2025let}, as it is the only open source method for multi‑person talking face generation.  Our method outperforms MultiTalk on nearly all metrics except SID, likely due to the stronger generalization ability of large video‑generation models. Moreover, our approach achieves real‑time FPS, while MultiTalk requires lengthy inference.

\begin{figure}[t!]
	\centering
	\includegraphics[width=\linewidth]{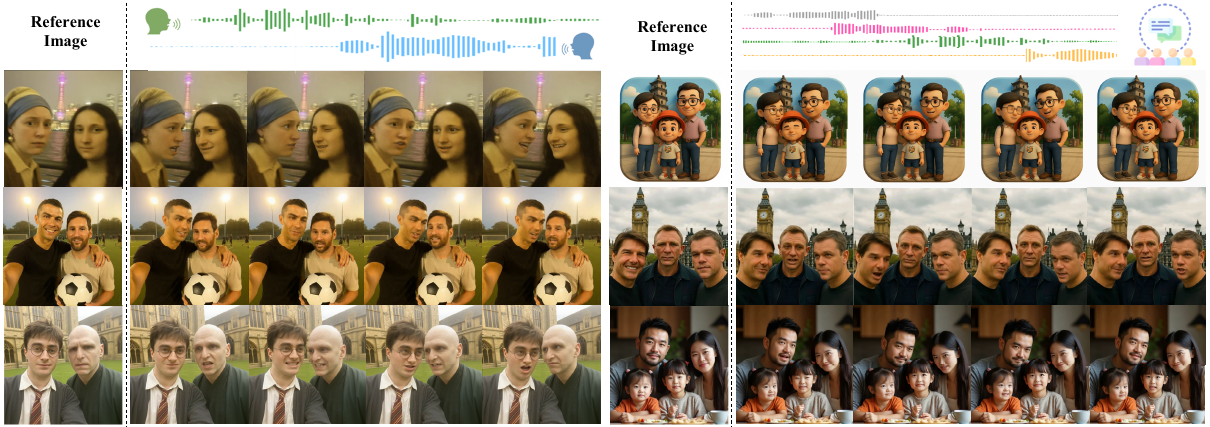}
	\vspace{-4mm} 
	\caption{\textbf{InterTalk for Multi‑person Conversation.}
				Our method supports diverse input styles and imposes no constraints on the number of participants, while producing highly interactive and realistic conversational behaviors.}
	\label{fig:result}
	\vspace{-3mm} 
\end{figure}

As shown in Fig.~\ref{fig:result}, our method not only generates natural and engaging interactions between participants but also produces multi‑participant conversational videos without any restricted on the number of participants. Additional visual comparisons with MultiTalk are provided in the supplementary material. In summary, our method demonstrates strong interactive capability and high efficiency in conversational talking‑face generation.

 \begin{table}[b]
\setlength{\tabcolsep}{4pt}
\footnotesize
\centering
 \vspace{-2mm}
\caption{Quantitative comparisons and ablation study on multi-person conversation generation on InterTalk Dataset.}
 \vspace{-1mm}
\resizebox{0.9\linewidth}{!}{

\begin{tabular}{c ccccc}
\toprule

    Methods & Sync-C $\uparrow$ & FD$\downarrow$ &  \text{RPCC}$\downarrow$ & \text{SID}$\uparrow$ & FPS$\uparrow$ \\ 

		\midrule
    MultiTalk \cite{kong2025let} & 7.94 & 41.74  & 0.24 & \textbf{5.14} &  0.61
    \\  

     \textbf{Ours}   & \textbf{8.30} & \textbf{22.67} & \textbf{0.09}  & 4.28 & \textbf{31.23}  \\

    \midrule
    \textit{w/o} Motion Feedback in RCE             & 8.13 & 29.72  & 0.15 & 2.71 & 31.85  \\
    \textit{w/o} Iterative Generation Strategy  & 8.25  & 24.26 & 0.11 & 4.12 &34.67  \\
    \textit{w/o} Disentanglement on Motion & 7.58  & 25.17 & 0.11 & 3.85 & 31.56  \\
    \textit{w/o} Data Augmentation  & 7.01  & 38.67 & 0.20 & 1.95 & 31.23  \\

\bottomrule
\end{tabular}
}

\label{table:main_table}
\end{table}

\subsection{Ablation Study}
As shown in Tab.~\ref{table:main_table}, we conduct an ablation study on four key components: the motion feedback in RCE, the iterative generation strategy, the disentanglement on motion refinement in IMG, and the 3D face‑based data augmentation. Removing motion feedback in RCE leads to a noticeable drop in interaction quality, as reflected by the interaction matrix on the FLAME coefficients. Introducing the iterative generation strategy significantly improves interaction performance with only a minor decrease in speed, as the model progressively refines motions to reduce response delays instead of relying solely on previously generated frames. The disentanglement allows the sync‑enhancer to further boost Sync‑C scores for lip synchronization by leveraging a large‑scale single‑speaker talking‑face dataset to compensate for the limited diversity of conversational data. Moreover, the 3D face‑based data augmentation greatly enhances interactive performance. Overall, the results show that each component contributes effectively to the framework.

As shown in Fig.~\ref{fig:iter}, we analyze how FD and SID vary with different iteration numbers in the iterative generation strategy. The results demonstrate that this strategy progressively improves both motion realism and diversity. Specifically, the FD steadily decreases with more iterations, indicating that the generated motion distribution becomes increasingly similar to real motion data. At the same time, the SID rises across iterations, demonstrating that the iterative refinement enhances motion variability and expressiveness. These results confirm the effectiveness of our iterative generation strategy in producing natural and diverse conversational behaviors.

\subsection{User Study}

\begin{figure}[t]
	\centering
	\begin{minipage}[t]{0.37\linewidth}
		\centering
		\includegraphics[width=\linewidth]{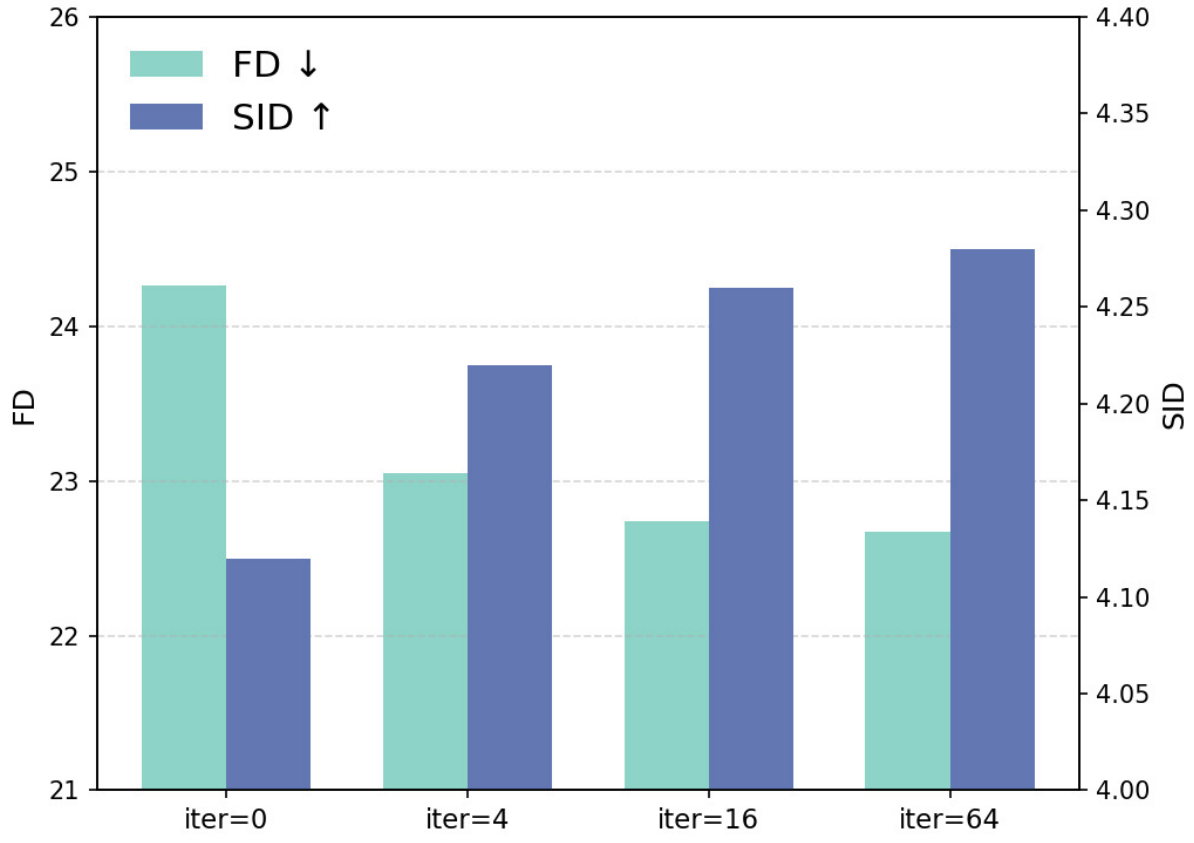}
		\vspace{-6mm}
		\caption{Variation of FD and SID across iterations.}
		\label{fig:iter}
	\end{minipage}
	\hfill
	\begin{minipage}[t]{0.58\linewidth}
		\centering
		\setlength{\tabcolsep}{4pt}
			\vspace{-32mm}
		\footnotesize
		\captionof{table}{User study on Interactive Head Generation. Best results are highlighted in bold.}
		\vspace{3mm}
		\resizebox{\linewidth}{!}{
			\begin{tabular}{c c c c c}
				\toprule
				\multirow{2}{*}{Methods} & Lip-sync & Interaction & Motion & Visual \\
				& Accuracy & Fluency & Naturalness & Realism \\
				\midrule
				DIM~\cite{tran2024dim} & 1.13 & 1.24 & 1.19 & 1.07 \\
				INFP~\cite{zhu2025infp} & 2.93 & 3.85 & 3.21 & 3.59 \\
				ARIG~\cite{guo2025arig} & 3.87 & 2.59 & 2.13 & 2.48 \\
				\textbf{Ours (Single)} & \textbf{4.03} & \textbf{3.96} & \textbf{4.55} & \textbf{4.12} \\
				\bottomrule
		\end{tabular}}
		\label{table:user_study}
	\end{minipage}
\vspace{-3mm}
\end{figure}

To evaluate perceptual quality, we conducted a user study on interactive head generation following the MOE protocol with 23 participants. As shown in Tab.~\ref{table:user_study}, each method was rated on four aspects: lip‑sync accuracy, interaction fluency, motion naturalness, and visual realism—using a 1–5 Likert scale where higher scores indicate better quality. Our method outperforms all baselines across every metric, demonstrating clear advantages in both realism and interactivity. Participants consistently preferred our results, praising the smoothness of facial motion and the natural transitions in interactive responses. In particular, our approach achieved the highest scores in motion naturalness and interaction fluency, verifying the effectiveness of our iterative generation strategy. Furthermore, the superior lip-sync accuracy highlights the strength of our sync enhancer module in producing temporally aligned mouth movements. These results collectively confirm that InterTalk delivers the most lifelike and engaging interactive head generation among all compared methods.

\section{Conclusion}

\label{sec:conclusion}
In this paper, we present InterTalk, a novel and efficient framework for highly interactive conversational talking‑face generation without constraints on the number of participants. The framework employs a Responsive Context Encoder (RCE) to encode audio and motion feedback into interactive features, ensuring flexibility. An Interactive Motion Generator (IMG) then synthesizes conversational motions with disentangled refinement, while a Rendering Pipeline animates multiple participants in real-time. We further incorporate an iterative generation strategy that refines motion feedback across participants, promoting coherent and responsive interactions. To address the scarcity of multi‑person conversational data, we construct a new dataset and introduce a 3D face‑based data augmentation strategy to enhance training. Extensive experiments demonstrate that InterTalk achieves flexible, natural, and efficient performance in interactive conversational talking face generation.

\section*{Acknowledgments}
\label{sec:acknowledge}
This work was supported in part by the New Generation Artificial Intelligence-National Science and Technology Major Project (No. 2025ZD0123501), Beijing Natural Science Foundation L242092, Chinese National Natural Science Foundation Projects 92570119, 62276254, U23B2054,  the Science and Technology Development Fund of Macau Project 0140/2024/AGJ, and InnoHK program.

\bibliographystyle{splncs04}
\bibliography{main}
\end{document}